\documentclass[11pt,a4paper]{article}
\usepackage[hyperref]{acl2020}
\usepackage{times}
\usepackage[utf8]{vietnam}
\usepackage{latexsym}

\aclfinalcopy % Uncomment this line for the final submission
\title{An Empirical Study of Using Pre-trained BERT Models for Vietnamese Relation Extraction Task at VLSP 2020}

\author{Pham Quang Nhat Minh \\
  Aimesoft JSC\\
  Hanoi, Vietnam\\
  {\tt minhpham@aimesoft.com}}

\date{}

\begin{document}
\maketitle
\begin{abstract}
In this paper, we present an empirical study of using pre-trained BERT models for the relation extraction task at the VLSP 2020 Evaluation Campaign. We applied two state-of-the-art BERT-based models: R-BERT and BERT model with entity starts. For each model, we compared two pre-trained BERT models: FPTAI/vibert and NlpHUST/vibert4news. We found that NlpHUST/vibert4news model significantly outperforms FPTAI/vibert for the Vietnamese relation extraction task. Finally, we proposed an ensemble model that combines R-BERT and BERT with entity starts. Our proposed ensemble model slightly improved against two single models on the development data and the test data provided by the task organizers.
\end{abstract}

\section{Introduction}
\label{sec:intro}

The relation extraction task is to extract entity mention pairs from a sentence and determine relation types between them. Relation extraction systems can be applied in question answering~\cite{xu2016question}, detecting contradiction~\cite{pham2013using}, and extracting gene-disease relationships~\cite{chun2006extraction}, protein-protein interaction~\cite{huang2004discovering} from biomedical texts.

In VLSP 2020, the relation extraction task is organized to assess and advance relation extraction work for the Vietnamese language. In this paper, we present an empirical study of BERT-based models for the relation extraction task in VLSP 2020. We applied two state-of-the-art BERT-based models for relation extraction: R-BERT~\cite{wu2019enriching} and BERT with entity starts~\cite{soares2019matching}. Two models use entity markers to capture location information of entity mentions. For each model, we investigated the effect of choosing pre-train BERT models in the task, by comparing two Vietnamese pre-trained BERT models: NlpHUST/vibert4news and FPTAI/vibert~\cite{the2020improving}. In our understanding, our paper is the first work that provides the comparison of pre-trained BERT models for Vietnamese relation extraction.

The remainder of this paper is structured as follows. In Section~\ref{sec:bert_model}, we present two existing BERT-based models for relation classification, which we investigated in our work. In Section~\ref{sec:method}, we describe how we prepared datasets for the two BERT-based models and our proposed ensemble model. In Section~\ref{sec:exp}, we give detailed settings and experimental results. Section~\ref{sec:discuss} gives discussions and findings. Finally, in Section~\ref{sec:conclusion}, we present conclusions and future work.

\section{BERT-based Models for Relation Classification}
\label{sec:bert_model}

In the following sections, we briefly describe BERT model~\cite{devlin-etal-2019-bert}, problem formalization, and two existing BERT-based models for relation classification, which we investigated in this paper.

\subsection{Pre-trained BERT Models}

The pre-trained BERT model~\cite{devlin-etal-2019-bert} is a masked language model that is built from multiple layers of bidirectional Transformer encoders~\cite{vaswani2017attention}. We can fine-tune pre-trained BERT models to obtain the state-of-the-art results on many NLP tasks such as text classification, named-entity recognition, question answering, natural language inference.

Currently, pre-trained BERT models are available for many languages. For Vietnamese, in our understanding, there are three available pre-trained BERT models: PhoBERT~\cite{phobert}, FPTAI/vibert~\cite{the2020improving}, and NlpHUST/vibert4news\footnote{vibert4news is available on https://huggingface.co/NlpHUST/vibert4news-base-cased}. Those models are different in pre-training data, selected tokenization, and training settings. In this paper, we investigated two pre-trained BERT models including FPTAI/vibert and NlpHUST/vibert4news for the relation extraction task. Investigation of PhoBERT for the task is left for future work.

\subsection{Problem Formalization}

In this paper, we focus on the relation classification task in the supervised setting. Training data is a sequence of examples. Each sample is a tuple $\mathbf{r}=(\mathbf{x},\mathbf{s}_1,\mathbf{s}_2,y)$. We define $\mathbf{x}=[x_0...x_n]$ as a sequence of tokens, where $x_0=[\textnormal{CLS}]$ is a special start marker. Let $\mathbf{s}_1=(i,j)$ and $\mathbf{s}_2=(k,l)$ are pairs of integers such that $0 < i \leq j \leq n, 0 < k \leq l \leq n$. Indexes of $\mathbf{s}_1$ and $\mathbf{s}_2$ are start and end indexes of two entity mentions in $\mathbf{x}$, respectively. $y$ denotes the relation label of the two entity mentions in the sequence $\mathbf{x}$. We use a special label $\textnormal{OTHER}$ for entity mentions which have no relation between them. Our task is to train a classification model from the training data.

\subsection{R-BERT}

In R-BERT~\cite{wu2019enriching}, for a sequence $\mathbf{x}$ and two target entities $e_1$ and $e_2$ which specified by indexes of $\mathbf{s}_1$ and $\mathbf{s}_2$, to make the BERT module capture the location information of the two entities, a special token '\$' is added at both the beginning and end of the first entity, and a special token '\#' is added at both the beginning and end of the second entity. [CLS] token is also added to the beginning of the sequence. 

For example, after inserting special tokens, a sequence with two target entities ``Phi Sơn'' and ``SLNA'' becomes to:

``[CLS] Cầu thủ \$ Phi Sơn \$ đã ghi bàn cho \# SLNA \# vào phút thứ 80 của trận đấu .''

The sequence $\mathbf{x}$ with entity markers, is put to a BERT model to get hidden states of tokens in the sequence. Then, we calculate averages of hidden states of tokens within the two target entities and put them through a tanh activation function and a fully connected layer to make vector representations of the two entities. Let $H'_0$, $H'_1$, $H'_2$ be hidden states at [CLS] and vector representations of $e_1$ and $e2$. We concatenate three hidden states and add a softmax layer for relation classification. R-BERT obtained 89.25\% of MACRO F1 on the SemEval-2010 Task 8 dataset~\cite{hendrickx-etal-2010-semeval}. 

\subsection{BERT with Entity Start}

We applied the BERT model with entity starts (hereinafter, referred to as BERT-ES) presented in~\cite{soares2019matching} for Vietnamese relation classification. In the model, similar to R-BERT, special tokens are added at the beginning and end of two target entities. In experiments of BERT-ES for Vietnamese relation classification, different from~\cite{soares2019matching}, we used entity markers `\$' and `\#' instead of markers `[E1]', `[/E1]', `[E1]', and `[/E2]'. We did not add [SEP] at the end of a sequence. In BERT-ES, hidden states at the start positions of two target entities are concatenated and put through a softmax layer for final classification. On SemEval-2010 Task 8 dataset, BERT-ES obtained 89.2\% of MACRO F1.

\section{Proposed Methods}
\label{sec:method}

In this work, we applied R-BERT and BERT-ES as we presented in Section~\ref{sec:bert_model} for Vietnamese relation extraction, and proposed an ensemble model of R-BERT and BERT-ES. In the following sections, we present how we prepared data for training BERT-based models and how we combined two single models: R-BERT and BERT-ES.

\subsection{Data Preprocessing}

Relation extraction data provided by VLSP 2020 organizers in WebAnno TSV 3.2 format~\cite{eckart-de-castilho-etal-2016-web}. In the data, sentences are not segmented and tokens are tokenized by white spaces. Punctuations are still attached in tokens. 

According to the task guideline, we consider only intra-sentential relations, so sentence segmentation is required in data preprocessing. We used VnCoreNLP toolkit~\cite{vu-etal-2018-vncorenlp} for both sentence segmentation and tokenization. For the sake of simplicity, we just used syllables as tokens of sentences. VnCoreNLP sometimes made mistakes in sentence segmentation, and as the result, we missed some relations for those cases.

\subsection{Relation Sample Generation}

\begin{table*}[t!]
\centering
\begin{tabular}{llll}
\hline \textbf{No.} & \textbf{Relation} & \textbf{Arguments} & \textbf{Directionality} \\ \hline
1 & LOCATED & PER - LOC, ORG – LOC & Directed\\
2 & PART–WHOLE & LOC – LOC, ORG – ORG, ORG-LOC & Directed\\
3 & PERSONAL–SOCIAL & PER – PER & Undirected\\
4 & AFFILIATION & PER – ORG, PER-LOC, ORG – ORG, LOC-ORG & Directed\\
\hline
\end{tabular}
\caption{\label{tbl:entityType} Relation types permitted arguments and directionality.}
\end{table*}

From each sentence, for training and evaluation, we made relation samples which are tupes $\mathbf{r}=(\mathbf{x},\mathbf{s}_1,\mathbf{s}_2,y)$ as described in Section~\ref{sec:bert_model}. Since in the data, named entities with their labels are provided, a simple way of making relation samples is generating all possible entity mention pairs from entity mentions of a sentence. We used the label OTHER for entity mention pairs that lack relation between them. All entity mentions pairs that are not included in gold-standard data are used as OTHER samples.

In the annotation guideline provided by VLSP 2020 organizers, there are constraints about types of two target entities of relation types as shown in Table~\ref{tbl:entityType}. Thus, we consider only entity mention pairs whose types satisfy those constraints. In training data, sometimes types of two target entities do not follow the annotation guideline. We accepted those entity pairs in making relation samples from provided train and development datasets. However, in processing test data for making submitted results, we consider only entity pairs whose types follow the annotation guideline.

Since the relation PERSONAL-SOCIAL is undirected, for this type, if we consider both pairs ($e_1$, $e_2$) and ($e_2$, $e_1$) in which $e_1$ and $e_2$ are PERSON entities, it may introduce redundancy. Thus, we added an extra constraint for PER-PER pairs that $e_1$ must come before $e_2$ in a sentence.

In the training data, we found a very long sentence with more than 200 relations. We omitted that sentence from the training data because that sentence may lead to too many OTHER relation samples.

\subsection{Proposed Ensemble Model}

In our work, we tried to combine R-BERT and BERT-ES to make an ensemble model. We did that by calculating weighted averages of probabilities returned by R-BERT and BERT-ES. Since in our experiments, BERT-ES performed slightly better than R-BERT on the development set, we used weights 0.4 and 0.6 for R-BERT and BERT-ES, respectively.

\section{Experiments and Results}
\label{sec:exp}

We conducted experiments to compare three BERT-based models on Vietnamese relation extraction data: R-BERT, BERT-ES, and the proposed ensemble model. We also investigated the effects of two Vietnamese pre-trained BERT models on the performance of models.

\subsection{Data}

\begin{table}[t!]
\centering
\begin{tabular}{lll}
\hline \textbf{Relation} & \textbf{Train} & \textbf{Dev} \\ \hline
LOCATED & 507 & 304 \\
PART-WHOLE & 1,016 & 402 \\
PERSONAL-SOCIAL & 101 & 95 \\
AFFILIATION & 756 & 489 \\
OTHER & 23,904 & 13,239 \\
\hline
Total & 26,284 & 14,529\\
\hline
\end{tabular}
\caption{\label{tbl:re_data} Label distribution of relation samples generated from train and dev data.}
\end{table}

The provided training dataset contains 506 documents, and the development dataset contains 250 documents. After data preprocessing and relation sample generation, we obtained relations with label distributions shown in Table~\ref{tbl:re_data}.

\subsection{Experimental Settings}

\begin{table}[t!]
\centering
\begin{tabular}{cc}
\hline \textbf{Hyper-Parameters} & \textbf{Value} \\ \hline
Max sequence length & 384 \\
Training epochs & 10 \\
Train batch size & 16\\
Learning rate & 2e-5\\
\hline
\end{tabular}
\caption{\label{tbl:parameter} Hyper-parameters used in training models.}
\end{table}

\begin{table*}[t!]
\centering
\begin{tabular}{llcc}
\hline \textbf{Model} & \textbf{Pre-trained BERT Model} & \textbf{MACRO F1} & \textbf{MICRO F1}\\ \hline
R-BERT & NlpHUST/vibert4news & 0.6392 & 0.7092 \\
R-BERT & FPTAI/vibert & 0.596 & 0.6736 \\
BERT-ES & NlpHUST/vibert4news & \textbf{0.6439} & 0.7101 \\
BERT-ES & FPTAI/vibert & 0.5976 & 0.6822 \\
Ensemble Model & NlpHUST/vibert4news & 0.6412 & \textbf{0.7108} \\
Ensemble Model & FPTAI/vibert & 0.6029 & 0.6851 \\
\hline
\end{tabular}
\caption{\label{tbl:result} Evaluation results on dev dataset.}
\end{table*}

\begin{table*}[t!]
\centering
\begin{tabular}{lcc}
\hline \textbf{Model} & \textbf{MACRO F1} & \textbf{MICRO F1}\\ \hline
R-BERT  & 0.6294 & 0.6645 \\
BERT-ES & 0.6276 & 0.6696 \\
Ensemble Model & \textbf{0.6342} & \textbf{0.6756} \\
\hline
\end{tabular}
\caption{\label{tbl:test_result} Evaluation results on test dataset.}
\end{table*}

In development, we trained models on the training data and evaluated models on the development data. However, to generate results on the provided test dataset, we trained BERT-based models on the dataset obtained by combining the provided training dataset and the development dataset.

Table~\ref{tbl:parameter} shows hyper-parameters we used for training models. We trained all models on a single 2080 Ti GPU.

We used MICRO F1 and MACRO F1 of four relation labels which do not include the label OTHER as evaluation measures.

\subsection{Results}

%\begin{table*}[t!]
%\centering
%\begin{tabular}{llcc}
%\hline \textbf{Model} & \textbf{Pre-trained BERT Model} & \textbf{MACRO F1} & \textbf{MICRO F1}\\ \hline
%R-BERT & NlpHUST/vibert4news & 0.6392 & 0.7092 \\
%R-BERT & FPTAI/vibert & 0.596 & 0.6736 \\
%BERT-ES & NlpHUST/vibert4news & \textbf{0.6439} & 0.7101 \\
%BERT-ES & FPTAI/vibert & 0.5976 & 0.6822 \\
%Ensemble Model & NlpHUST/vibert4news & 0.6412 & \textbf{0.7108} \\
%Ensemble Model & FPTAI/vibert & 0.6029 & 0.6851 \\
%\hline
%\end{tabular}
%\caption{\label{tbl:result} Evaluation results on dev dataset.}
%\end{table*}
%
%\begin{table*}[t!]
%\centering
%\begin{tabular}{lcc}
%\hline \textbf{Model} & \textbf{MACRO F1} & \textbf{MICRO F1}\\ \hline
%R-BERT  & 0.6294 & 0.6645 \\
%BERT-ES & 0.6276 & 0.6696 \\
%Ensemble Model & \textbf{0.6342} & \textbf{0.6756} \\
%\hline
%\end{tabular}
%\caption{\label{tbl:test_result} Evaluation results on test dataset.}
%\end{table*}

Table~\ref{tbl:result} shows the evaluation results obtained on the development dataset. We can see that using NlpHUST/vibert4news  significantly outperformed FPTAI/vibert in both MICRO F1 and MACRO F1 scores. BERT-ES performed slightly better than R-BERT. The proposed ensemble model is slightly improved against R-BERT and BERT-ES in terms of MICRO F1 score.

Table~\ref{tbl:test_result} shows the evaluation results obtained on the test dataset. We used NlpHUST/vibert4news for generating test results. Table~\ref{tbl:test_result} confirmed the effectiveness of our proposed ensemble model. The ensemble model obtained the best MACRO F1 and the best MICRO F1 score on the test data among the three models.

\subsection{Result Analysis}

\begin{table*}[t!]
\centering
\begin{tabular}{llcc}
\hline & \textbf{Precision} & \textbf{Recall} & \textbf{F1}\\ \hline
AFFILIATION & 0.7615 & 0.744 & 0.7528 \\
LOCATED & 0.7053 & 0.7007 & 0.7030 \\
PART – WHOLE & 0.65 & 0.8085 & 0.7206\\
PERSONAL - SOCIAL & 0.6136 & 0.2842 & 0.3885\\
\hline
\end{tabular}
\caption{\label{tbl:analysis} Precision, Recall, F1 for each relation type on the dev dataset.}
\end{table*}

We looked at details of precision, recall, and F1 scores for each relation type on the development data. Table~\ref{tbl:analysis} shows results of the ensemble model with vibert4news pre-trained model. PERSONAL-SOCIAL turned out to be a difficult label. The proposed ensemble obtained a low Recall, and F1 score for that label. The reason might be that the relations of PERSONAL-SOCIAL are few in the training data while the patterns of PERSONAL-SOCIAL relations are wider than other relation types.

\section{Discussion}
\label{sec:discuss}

\begin{table}[t!]
\centering
\begin{tabular}{ccc}
\hline & \textbf{FPTAI/vibert} & \textbf{vibert4news}\\ \hline
Data size & 10GB & 20GB\\
Data domain & News & News\\
Tokenization & Subword & Syllable\\
Vocab size & 38168 & 62000\\
\hline
\end{tabular}
\caption{\label{tbl:pretrain_bert} Comparison of NlpHUST/vibert4news and FPTAI/vibert.}
\end{table}

In experiments, we compared the effects of two pre-trained BERT models: NlpHUST/vibert4news and FPTAI/vibert on relation extraction. The two pre-trained models have the same BERT architecture (BERT base model) but are different in chosen tokenizers, vocabulary size, pre-training data, and training procedure. Table~\ref{tbl:pretrain_bert} shows a comparison of the two models.

FPTAI/vibert was trained on 10GB of texts collected from online newspapers while NlpHUST/vibert4news was trained on 20GB of texts in the news domain. FPTAI/vibert used subword tokenization, and vocabulay of FPTAI/vibert was modified from mBERT while tokenization of vibert4news is based on syllables.

We come up with some reasons why using NlpHUST/vibert4news significantly outperformed FPTAI/vibert for Vietnamese relation extraction.

\begin{itemize}
    \item Pre-training data used to trained vibert4news is much larger than FPTAI/vibert.
    \item Tokenization used in NlpHUST/vibert4news is based on syllables while FPTAI/vibert used subwords and modified the original vocabulary of mBERT. We hypothesize that syllables which are basic units in Vietnamese are more appropriate than subwords for Vietnamese NLP tasks.
\end{itemize}

Due to the time limit, we did not investigate PhoBERT~\cite{phobert} which used word-level corpus to train the model. As future work, we plan to compare vibert4news that uses syllable-based tokenization with PhoBERT that uses word-level/subword tokenization for Vietnamese relation extraction.

\section{Conclusion}
\label{sec:conclusion}

We have presented an empirical study of BERT-based models for relation extraction task at VLSP 2020 Evaluation Campaign. Experimental results show that the BERT-ES model which uses entity markers and entity starts obtained better results than the R-BERT model, and choosing an appropriate pre-trained BERT model is important for the task. We showed that pre-trained model NlpHUST/vibert4news outperformed FPTAI/vibert for Vietnamese relation extraction task. In future work, we plan to investigate PhoBERT~\cite{phobert} for Vietnamese relation extraction to understand the effect of using word segmentation to the task.

\bibliography{vlsp2020}
\bibliographystyle{acl_natbib}

\end{document}